\documentclass[conference]{IEEEtran}
\IEEEoverridecommandlockouts
\usepackage{cite}
\usepackage{amsmath,amssymb,amsfonts}
\usepackage{algorithmic}
\usepackage{graphicx}
\usepackage{textcomp}
\usepackage{xcolor}

\usepackage{times}
\usepackage{epsfig}
\usepackage{pifont}

\usepackage{mathtools}
\usepackage{booktabs}
\usepackage{multirow}
\usepackage{adjustbox}
\usepackage{rotating}
\usepackage{paralist}
\usepackage{comment}
\usepackage{bbm}
\usepackage{colortbl}
\usepackage{pbox}
\usepackage{xspace}
\usepackage{hyperref}

\makeatletter
\DeclareRobustCommand\onedot{\futurelet\@let@token\@onedot}
\def\@onedot{\ifx\@let@token.\else.\null\fi\xspace}
 
\def\ie{\emph{i.e}\onedot}

\makeatletter

\def\BibTeX{{\rm B\kern-.05em{\sc i\kern-.025em b}\kern-.08em
    T\kern-.1667em\lower.7ex\hbox{E}\kern-.125emX}}
\begin{document}

\title{Reimagine BiSeNet for Real-Time Domain Adaptation in Semantic Segmentation

}

\author{\IEEEauthorblockN{Antonio Tavera}
\IEEEauthorblockA{\textit{DAUIN}
\textit{Politecnico di Torino}\\
antonio.tavera@polito.it}
\and
\IEEEauthorblockN{Carlo Masone}
\IEEEauthorblockA{\textit{CINI},
\textit{Politecnico di Torino}\\
}
\and
\IEEEauthorblockN{Barbara Caputo}
\IEEEauthorblockA{\textit{DAUIN}
\textit{Politecnico di Torino}\\
barbara.caputo@polito.it}
}

\maketitle

\begin{abstract}
Semantic segmentation models have reached remarkable performance across various tasks. However, this performance is achieved with extremely large models, using powerful computational resources and without considering training and inference time.
Real-world applications, on the other hand, necessitate models with minimal memory demands, efficient inference speed, and executable with low-resources embedded devices, such as self-driving vehicles.

In this paper, we look at the challenge of real-time semantic segmentation across domains, and we train a model to act appropriately on real-world data even though it was trained on a synthetic realm. We employ a new lightweight and shallow discriminator that was specifically created for this purpose. To the best of our knowledge, we are the first to present a real-time adversarial approach for assessing the domain adaption problem in semantic segmentation. We tested our framework in the two standard protocol: GTA5$\to$Cityscapes and SYNTHIA$\to$Cityscapes. Code is available at: \href{https://github.com/taveraantonio/RTDA}{https://github.com/taveraantonio/RTDA}

\end{abstract}

\section{Introduction}
Semantic segmentation, \ie, assigning a semantic class to each pixel of an image, is a critical task for scene comprehension. It is fraught with challenges and the state-of-the-art models proposed to tackle them usually have a huge number of parameters. The complexity of these models not only translates to long training and inference times but it also makes it impractical to deploy them in a real-world scenario due to the large amount of resources demanded. 
Moreover, semantic segmentation id often required to work in real-time, particularly for robotics applications such as geo sensing, precision agriculture, and, most notably, autonomous driving.

Besides the complexity of the models, the process of collecting and annotating real-world data \cite{Cordts2016Cityscapes} is time-consuming and costly. A successful solution to tackle this issue is to use synthetic data generated from virtual world simulators \cite{Richter2016GTA}, \cite{Ros2016Synthia}, \cite{tavera2020idda}. Despite the much lower cost of collecting and annotating synthetic data, this technique has one major drawback: the domain shift between virtual and real world is substantial.
Several unsupervised domain adaptation techniques have been proposed to address the domain gap between the synthetic (\textit{source}) and real (\textit{target}) domains; however, because they are not designed to be used in a real-world scenario and rely on a huge number of parameters, they are still  vulnerable to resource and training time limits.

To fully solve the real-time domain adaptation problem in semantic segmentation, we require a complete lightweight model with few parameters and that can be deployed in a practical situation with limited resources.
To do this, we redesigned the BiSeNet \cite{bisenetv1} model, tailoring it to the Domain Adaptation challenge and including a novel lighter and thinner fully convolutional domain discriminator (Light\&Thin). To summarize: 

\begin{itemize}
    \item we propose a network for real-time domain adaptation in semantic segmentation, using a new lightweight and thin domain discriminator.
    \item we propose an ablation study to compare our Light\&Thin discriminator to a standard domain discriminator and its lightweight variant.
    \item we test our architecture against two synthetic-to-real situations, GTA$\to$Cityscapes and Synthia$\to$Cityscapes, proving the efficacy of our solution. 
\end{itemize}

\begin{figure*}[t]
\begin{center}
\includegraphics[width=1.0\textwidth]{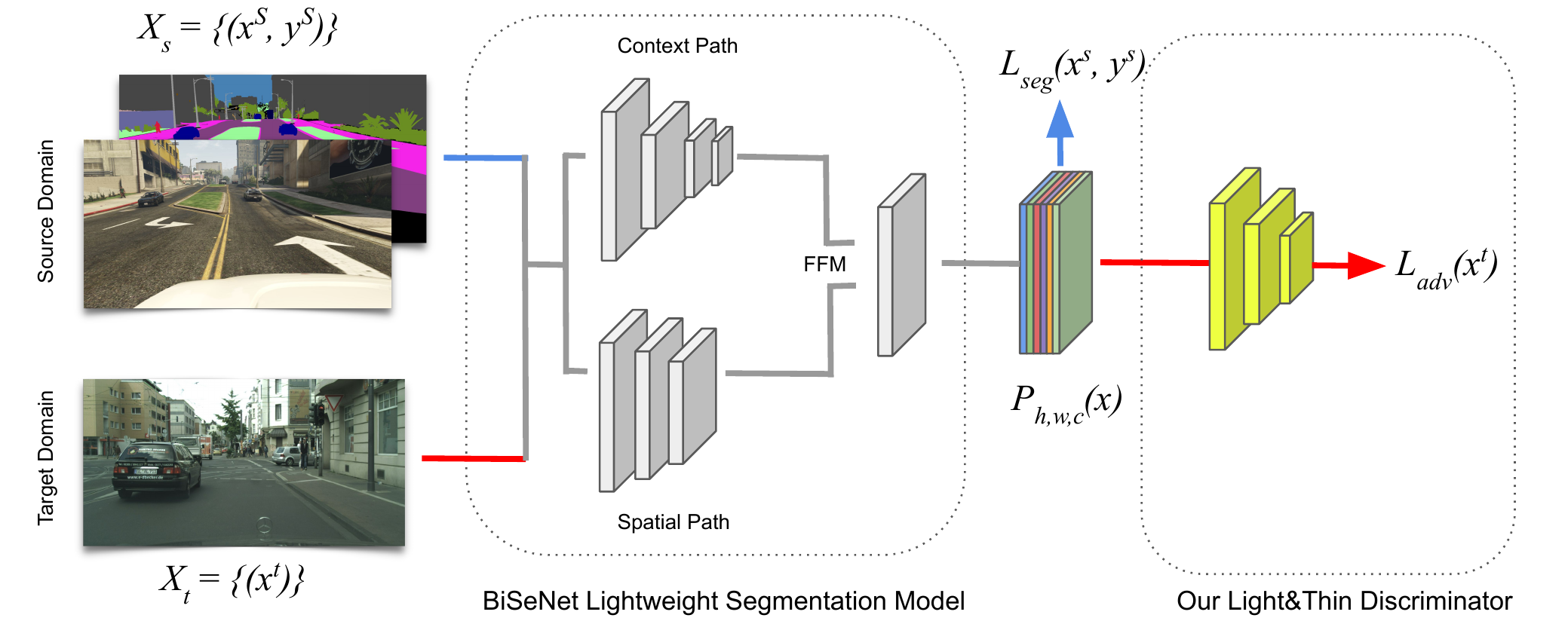}
\end{center}
\vspace{-10pt}
\caption{
Illustration of the real-time adversarial training of our framework. The adversarial loss required to align the source and target distributions is computed by our novel lightweight and shallow discriminator.
}
\vspace{-10pt}
\label{fig:method}
\end{figure*}

\section{Related Works}

\subsection{Semantic Segmentation and real-time application}
Thanks to the use of deep learning techniques, Semantic Segmentation has exploded in popularity in recent years.
The current state-of-the-art methods are determined by the approach employed to exploit semantic information, such as fully convolutional networks \cite{long2015fully}, encoder-decoder architectures \cite{noh2015learning}, \cite{unet}, dilated convolutions \cite{chen2017deeplab}, \cite{chen2018encoder}, \cite{chen2017rethinking} or multi-scale and pyramid networks \cite{zhao2017pyramid}.
Because the number of parameters in semantic segmentation networks is in the order of $10^9$ and their real-world application is rising in popularity, several researchers have investigated the feasibility of more lightweight architectures. The majority of architectures can be divided into two macro categories: (i) encoder-decoder architectures \cite{mehta2018espnet}, \cite{erfnet}, \cite{Zhang2018ShuffleNetAE}, which cost less at inference time than dilated convolution methods, (ii) two-pathway architectures, which address the loss of semantic information during the encoder-decoder mechanism's downsampling and upsampling operations. The BiSeNet family \cite{bisenetv1}, \cite{bisenetv2} is an example of this type of architecture.

\subsection{Domain Adaptation}
The task of bridging the gap between two different distributions is referred to as Domain Adaptation. The original answer to this problem is to employ a distance minimization algorithm, such as the MMD \cite{geng2011daml}, although alternative methods that use generative models \cite{li2019bidirectional}, \cite{kim2020learning} to condition one domain into the other have also been used. The most noteworthy solution is the adversarial training technique \cite{chang2019all}, \cite{vu2019advent}, which consists in a min-max game between the segmentation network and a discriminator in which the former attempts to trick the latter by making the distributions of the two domains identical. In any case, none of the prior solutions were applicable to real-world scenario.

\section{Method}
The proposed algorithm re-imagines BiSeNet tailored to the Unsupervised Domain Adaptation (UDA) task (\ref{sec:setting}). We introduce a novel real-time adversarial domain adaption framework (\ref{sec:training}) comprised of the BiSeNet semantic segmentation model and an unique lightweight and thin discriminator (Fig. \ref{fig:method}) that increases domain alignment and adaptation performance. 

\subsection{Setting}
\label{sec:setting}
The set of RGB images composed by $\mathcal{I}$ pixels is denoted by $\mathcal{X}$, and the set of semantic labels linking each pixel I with a class $c$ from a set of semantic classes $\mathcal{C}$ is denoted by $\mathcal{Y}$. We have two datasets to work with during training: the source $X_{s} = \{(x^{s}, y^{s})\}$, which consists of $|X_{s}|$ semantically annotated images, and the target $X_{t} = \{(x^{t})\}$, which consists of $|X_{t}|$ unlabeled images. The source and target annotation mask belonging to the set of semantic labels $\mathcal{Y}$ are defined as $y^{s}$ and $y^{t}$. The goal of UDA is to use both the source and target dataset  $X_{s}$ and $X_{t}$ to learn a function $f$ that takes as input an image $x$ and outputs a $C$-dimensional segmentation map $P_{h,w,c}(x)$.

\subsection{Training}
\label{sec:training}
Due to the lack of semantic information for the target distribution, we proceed to align the features derived from the source and target domains in an adversarial fashion. To do this, as well as to meet our goal of making a network smaller, portable, and deployable on limited resource devices, we require a different domain discriminator. This is why we developed and tested two different types of lighter discriminators: a less expensive version ($D_{Light}$) of the widely used Fully Convolutional discriminator \cite{radford2016unsupervised} and a shallow version ($D_{Light\&Thin}$) of the latter. Both discriminators $D$ employ depthwise separable convolution instead of the conventional convolution and are trained to discriminate between source and target domains using the following loss:
\begin{equation}
    L_{D}(x^s, x^t) = - \sum_{h,w}
     \log\ D(P^s_{h,w,c}) +
     \log(1-D(P^t_{h,w,c})).
     \label{eq:loss_discriminator}
\end{equation}
More details on these two lightweight discriminators are presented in Sec.~\ref{sec:experiments}.

The adversarial training is carried out using the features extracted by the semantic segmentation model and the domain prediction coming from a discriminator model. Both models engage in a min-max game in which the discriminator guesses the domain to which a feature belongs to and the segmentation network attempts to mislead the discriminator by making features from both domains similar. To accomplish this effect an adversarial loss $L_{adv}$ is used as follow: 
\begin{equation}
    L_{\text{adv}}(x^t) = - \frac{1}{|X_t|} \sum_{h,w} \log D(P^t_{h,w,c}).
    \label{eq:adversarial_loss}
\end{equation}

\begin{table*}[t]
\begin{adjustbox}{width=1.0\textwidth}
\centering

\begin{tabular}{l|lllllllllllllllllll|l}
Experiment    & \begin{sideways}Road\end{sideways} & \begin{sideways}Sidewalk\end{sideways} & \begin{sideways}Building\end{sideways} & \begin{sideways}Wall\end{sideways} & \begin{sideways}Fence\end{sideways} & \begin{sideways}Pole\end{sideways} & \begin{sideways}TLight\end{sideways} & \begin{sideways}TSign\end{sideways} & \begin{sideways}Vegetation\end{sideways} & \begin{sideways}Terrain\end{sideways} & \begin{sideways}Sky\end{sideways} & \begin{sideways}Person\end{sideways} & \begin{sideways}Rider\end{sideways} & \begin{sideways}Car\end{sideways} & \begin{sideways}Truck\end{sideways} & \begin{sideways}Bus\end{sideways} & \begin{sideways}Train\end{sideways} & \begin{sideways}Motorcycle\end{sideways} & \begin{sideways}Bicycle\end{sideways} & mIoU$^{19}$      \\ 
\hline
Target Only   & 97.11                              & 77.88                                  & 88.67                                  & 48,31                              & 48.31                               & 41.33                              & 39.56                                & 48.36                               & 58.27                                    & 89.07                                 & 57.03                             & 91.86                                & 66.45                               & 40.47                             & 90.63                               & 60.23                             & 67.11                               & 50.32                                    & 44.93                                 & 64.17            \\ 
\hline
FCD           & 74.21                              & 28.66                                  & 72.47                                  & 12.57                              & 16.31                               & 12.72                              & 28.03                                & 17.32                               & 80.17                                    & 14.64                                 & 77.31                             & 41.97                                & 8.88                                & 65.58                             & 18.2                                & 6.85                              & 18.33                               & 11.71                                    & 0.0                                   & 31.89            \\
FCD-Light     & 83.17                              & 33.53                                  & 68.9                                   & 11.37                              & 7.59                                & 13.46                              & 25.12                                & 14.51                               & 79.49                                    & 30.09                                 & 74.97                             & 41.47                                & 13.61                               & 67.73                             & 19.84                               & 7.05                              & 4.92                                & 14.63                                    & 0.0                                   & 32.18            \\
FCD-LightThin & 83.92                              & 37.21                                  & 74.23                                  & 14.19                              & 15.63                               & 17.61                              & 29.93                                & 19.16                               & 79.85                                    & 24.91                                 & 72.14                             & 43.24                                & 11.15                               & 61.0                              & 17.41                               & 14.28                             & 7.16                                & 8.22                                     & 0.0                                   & \textbf{33.22 } 
\end{tabular}

\end{adjustbox}
\vspace{1pt}
\caption{GTA5$\to$Cityscapes Unsupervised Domain Adaptation experiments. FCD stands for Fully Convolutional Discriminator. FCD-Light indicates our lightweight variant. FCD-Light\&Thin indicates our thinner and lightweight discriminator.}
\vspace{-10pt}
\label{table:gta_exp}
\end{table*}
\begin{table*}[t]
\begin{adjustbox}{width=1.0\textwidth}
\centering

\begin{tabular}{l|llllllllllllllll|l}
Experiment    & \begin{sideways}Road\end{sideways} & \begin{sideways}Sidewalk\end{sideways} & \begin{sideways}Building\end{sideways} & \begin{sideways}Wall\end{sideways} & \begin{sideways}Fence\end{sideways} & \begin{sideways}Pole\end{sideways} & \begin{sideways}TLight\end{sideways} & \begin{sideways}TSign\end{sideways} & \begin{sideways}Vegetation\end{sideways} & \begin{sideways}Sky\end{sideways} & \begin{sideways}Person\end{sideways} & \begin{sideways}Rider\end{sideways} & \begin{sideways}Car\end{sideways} & \begin{sideways}Bus\end{sideways} & \begin{sideways}Motorcycle\end{sideways} & \begin{sideways}Bicycle\end{sideways} & mIoU$^{16}$      \\ 
\hline
Target Only   & 97.11                              & 77.88                                  & 88.67                                  & 48,31                              & 48.31                               & 41.33                              & 39.56                                & 48.36                               & 58.27                                    & 57.03                             & 91.86                                & 66.45                               & 40.47                             & 60.23                             & 50.32                                    & 44.93                                 & 64.17            \\ 
\hline
FCD           & 72.74                              & 32.18                                  & 75.31                                  & 4.45                               & 0.35                                & 14.07                              & 0.09                                 & 2.58                                & 66.39                                    & 80.87                             & 35.84                                & 3.32                                & 54.26                             & 18.08                             & 1.49                                     & 9.18                                  & 29.45            \\
FCD-Light     & 68.02                              & 33.17                                  & 74.76                                  & 8.69                               & 0.32                                & 16.41                              & 6.25                                 & 4.77                                & 56.92                                    & 80.67                             & 37.33                                & 4.64                                & 50.61                             & 17.7                              & 3.49                                     & 16.63                                 & 30.02            \\
FCD-LightThin & 63.01                              & 23.5                                   & 76.94                                  & 8.71                               & 0.74                                & 19.93                              & 9.04                                 & 7.52                                & 76.56                                    & 79.98                             & 44.01                                & 4.29                                & 63.76                             & 14.56                             & 1.99                                     & 11.97                                 & \textbf{31.66 } 
\end{tabular}

\end{adjustbox}
\vspace{1pt}
\caption{SYNTHIA$\to$Cityscapes Unsupervised Domain Adaptation experiments. FCD stands for Fully Convolutional Discriminator. FCD-Light indicates our lightweight variant. FCD-Light\&Thin indicates our thinner and lightweight discriminator.}
\vspace{-20pt}
\label{table:synthia_exp}
\end{table*}

We jointly optimize the supervised segmentation loss $L_{seg}$ on source samples and the unsupervised entropy loss $L_{adv}$ on target samples while training the BiSeNet semantic segmentation model. The following is the definition of the total loss function:
\begin{equation}
    \frac{1}{|X_s|} \sum_{(x^s,y^s) \in X_s} L_{\text{seg}}(x^s, y^s) + \frac{1}{|X_t|} \sum_{x^t \in X_t} + \lambda L_{adv}(x^t),
    \label{eq:total_loss}
\end{equation}
where $L_{seg}$ minimize the standard cross-entropy loss defined as:
\begin{equation}
    L_{\text{seg}}(x^s, y^s) = - \frac{1}{|X_s|} \sum_{h,w} \sum_c y_{h,w,c}^s \log(P^s_{h,w,c})
    \label{eq:crossentropy_loss}
\end{equation}


\section{Experiments}\label{sec:experiments}
\subsection{Datasets}
We test our model over the two standard synthetic-to-real benchmarks in Domain Adaptation for Semantic Segmentation: GTA5$\to$Cityscapes and SYNTHIA$\to$Cityscapes.

GTA5 \cite{Richter2016GTA} is made up of 24966 annotated photos from the aforementioned video-game. The standard 19-classes, which Cityscapes shares, is used for training and evaluation.

SYNTHIA \cite{Ros2016Synthia} is made up of 9400 annotated images from a virtual world and belonging to the RAND-CITYSCAPES subset.  The usual 19-classes shared by Cityscapes are utilized for training, whereas the assessment is performed on 16-classes using the \cite{vu2019advent} protocol.

Cityscapes \cite{Cordts2016Cityscapes}  is made up of 2975 real-world pictures gathered from various German cities. To test our network, we use the entire validation set of 500 photos at the original 2048x1024 resolution.

\subsection{Implementation details}
The segmentation model of our method is BiSeNet \cite{bisenetv1} with the Context Path (see section 3.2 of \cite{bisenetv1}) initialized with a ResNet-101\cite{he2016deep} pretrained on ImageNet. The standard discriminator used for the comparison is a common Fully Convolutional Discriminator (FCD) with 5 convolution layers with kernel size $4x4$, channel numbers $\{64, 128, 256, 512, 1\}$, padding 2 and stride 1. Its lightweight variant (FCD-Light) is obtained by substituting each convolution operation with a depthwise-separable convolution \cite{xception}, comprises of a depthwise convolution done independently over each input channels, followed by a pointwise convolution, with kernel size $1x1$. Our thinner version (FCD-Light\&Thin) has only 3 depthwise separable convolution layers with channel numbers $\{64, 128, 1\}$. Each convolution or depthwise separable convolution layer is followed by a Leaky ReLU with negative slope 0.2.

PyTorch is used to implement our technique. The segmentation model is trained with batch size $4$ and SGD with an initial learning rate of $2.5\times10-4$, which is then changed at each iteration with a "poly" learning rate decay with power $0.9$, momentum $0.9$, and weight decay $0.0005$. Adam is used to train all of the discriminators, with momentum $(0.9, 0.99)$, learning rate $10-5$, and the same segmentation model scheduler. The model has undergone $30k$ iterations of training. The value of $\lambda_{adv}$ is set to $0.01$. The training images are shrunk to $(1024,512)$, whereas the evaluation is done on the $(2048,1024)$ original image dimension. 

We use the standard Intersection over Union metric to measure the performance of our experiments.

\begin{figure*}[ht]
\begin{center}
\includegraphics[width=1.0\textwidth]{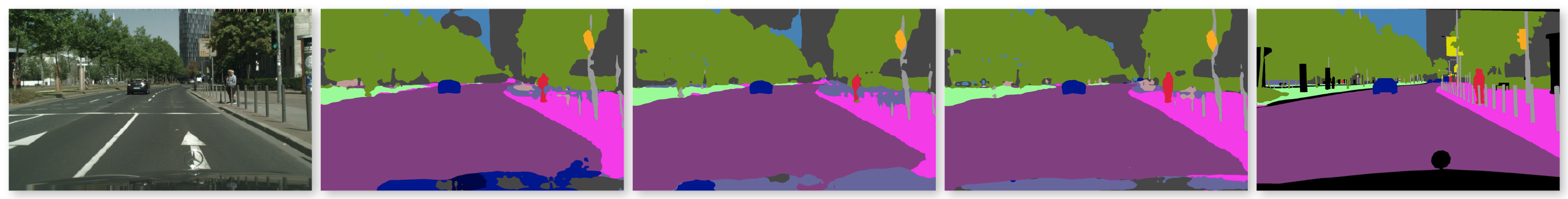}
\end{center}
\vspace{-10pt}
\caption{
Qualitative results for the GTA$\to$Cityscapes experiment. Starting from the left: RGB, FCD, FCD-Light, FCD-Light\&Thin, Ground Truth.
}
\vspace{-10pt}
\label{fig:qualitative}
\end{figure*}

\subsection{Results}
Table \ref{table:gta_exp} and Table \ref{table:synthia_exp} show the result on the GTA$\to$Cityscapes and SYNTHIA$\to$Cityscapes, respectively.
By looking at Table \ref{table:gta_exp}, it is clear that using a typical Fully Convolutional discriminator (FCD) we get performances that are approximately half of what we would achieve if we trained directly on the target. When each convolution in this discriminator is replaced with its lightweight counterpart (FCD-Light), we get comparable results with just a $+0,29\%$ gain in accuracy. However, as seen in Table \ref{table:parameters}, the number of parameters and FLOPS decreases significantly, as does training and inference time. When the input resolution is 1024x512, the difference in parameters is 2.59 million, while the FLOPS move from 30.883G to barely 2.14G. 
When we use our light and shallow discriminator (Light\&Thin), the reduction in parameters and FLOPS is proportionate to an enhancement in accuracy; indeed, our solution improves performance by $+1,33\%$ over the typical FCD. Since the task is classification, using a shallow domain discriminator like ours takes less epochs to attain a local optima than a conventional DCGAN discriminator, which would require more epochs and longer training time to converge.
We would want to emphasize that all of this results were collected while training on two TESLA v100 GPUs rather than on commercial hardware such as a Jetson Javier.
The SYNTHYA$\to$Cityscapes experiment described in Table \ref{table:synthia_exp} shows a similar pattern. Replacing common convolutions with depthwise separable convolutions results in a small $+0.57\%$ improvement, but when utilizing our Light\&Thin discriminator, an average boost of $+2,21\%$ is attained.
Figure \ref{fig:qualitative} confirms this tendency; as you can see, our Light\&Thin model allows for better segmentation, even for small classes like pedestrians, poles or traffic signs. It should be noted that these results come from models that were trained on synthetic data with a distribution that is substantially different from the real-world test set. There is still work to be done to improve performance and bridge the gap between the two domains and the existing state-of-the-art but non-real-time domain adaptation models.
\begin{table}[t]
\begin{adjustbox}{width=1.0\columnwidth}
\centering

\begin{tabular}{l|lll}
           & FCD     & FCD-Light & FCD-Light\&Thin  \\ 
\hline
Parameters & 2.781M  & 0.191M~   & 13.316K        \\
FLOPS      & 30.883G & 2.14G     & 1.038G        \\
Training Time & 7h:04m & 6h:39m & 6h:32m
\end{tabular}

\end{adjustbox}
\vspace{1pt}
\caption{Comparison between the number of parameters, FLOPS and Training Time among the three distinct discriminators used.}
\vspace{-20pt}
\label{table:parameters}
\end{table}

\vspace{-5pt}
\section{Conclusion}
In this paper, we look at Real Time Domain Adaptation in Semantic Segmentation. The primary goal is to minimize model parameters as well as training and inference time in order to make the model feasible for real-world applications. We present a whole lightweight framework that includes a unique light and shallow discriminator. We evaluated our approach using the two common synthetic-to-real protocols.
The results indicate that there is still work to be done in this task; future research will focus on applying our discriminator to more complex and powerful lightweight semantic segmentation models, as well as enhancing the entire framework. 

{\small
\bibliographystyle{unsrt}
\bibliography{egbib}
}
\end{document}